%% file: main.tex
\documentclass[10pt,twocolumn,letterpaper]{article}
\usepackage[pagenumbers]{cvpr}

\usepackage{subcaption}
\usepackage{tabularx}
\usepackage{graphicx}
\usepackage{amsmath}
\usepackage{amssymb}
\usepackage{booktabs}
\usepackage{hyperref}
\usepackage{xcolor}  
\usepackage{forest}

\usepackage{multirow}
\usepackage{colortbl}
\usepackage{float}
\usepackage{xcolor}  
\usepackage{graphicx}

\def\confName{CVPR}
\def\confYear{2026}

\title{Hierarchy-Aware Fine-Tuning of Vision-Language Models}

\author{
  Jiayu Li$^{1}$\thanks{Work done during Google Summer of Code 2025 with Intel: \url{https://summerofcode.withgoogle.com/archive/2025/projects/eYkkpBH5}} \quad
  Rajesh Gangireddy$^{2}$ \quad
  Samet Akcay$^{2}$ \quad
  Wei Cheng$^{1}$ \quad
  Juhua Hu$^{1}$ \\
  $^1$University of Washington \quad
  $^2$Intel\\
}

\begin{document}
\maketitle


\begin{abstract}
    Vision–Language Models (VLMs) learn powerful multimodal representations through large-scale image–text pretraining, but adapting them to hierarchical classification is underexplored. Standard approaches treat labels as flat categories and require full fine-tuning, which is expensive and produces inconsistent predictions across taxonomy levels. We propose an efficient hierarchy-aware fine-tuning framework that updates a few parameters while enforcing structural consistency. We combine two objectives: Tree-Path KL Divergence (TP-KL) aligns predictions along the ground-truth label path for vertical coherence, while Hierarchy-Sibling Smoothed Cross-Entropy (HiSCE) encourages consistent predictions among sibling classes. Both losses work in the VLM's shared embedding space and integrate with lightweight LoRA adaptation. Experiments across multiple benchmarks show consistent improvements in Full-Path Accuracy and Tree-based Inconsistency Error with minimal parameter overhead. Our approach provides an efficient strategy for adapting VLMs to structured taxonomies.
\end{abstract}

\input{content/introduction}
\input{content/related_work}
\input{content/method}
\input{content/experiment}

\input{content/conclusion}

\bibliographystyle{ieeenat_fullname}
\bibliography{ref}

\end{document}

%% file: content/introduction.tex
\section{Introduction}

Vision–Language Models (VLMs) unify visual perception and language understanding in a single multimodal framework. A typical VLM architecture combines a visual encoder, a language encoder or decoder, and a cross-modal alignment module that creates a shared embedding space for image–text reasoning. These architectures enable zero-shot recognition, visual question answering, and caption generation through large-scale image–text pretraining.

Pretraining provides broad generalization but doesn't ensure optimal performance on domain-specific tasks. Differences in label structure, data distribution, and visual characteristics require task-adaptive fine-tuning. Fully updating billions of parameters demands substantial GPU memory and training cost. Efficient Fine-Tuning (EFT) methods ~\cite{hu2023vl, prompt,lora, qlora,adapter} address this by optimizing a small subset of parameters while keeping pretrained weights frozen.
Recent studies show that EFT methods achieve strong performance on visual question answering ~\cite{liu2023parameter,van2023open,zhang2024cream}, image captioning~\cite{duan2024cityllava,wang2023efficientvlm}, cross-modal retrieval~\cite{lu2023uniadapter,yang2024cross} and text-guided object detection~\cite{madan2024revisiting,sharshar2025vision}.

\begin{figure}
    \centering
    \includegraphics[width=1\linewidth]{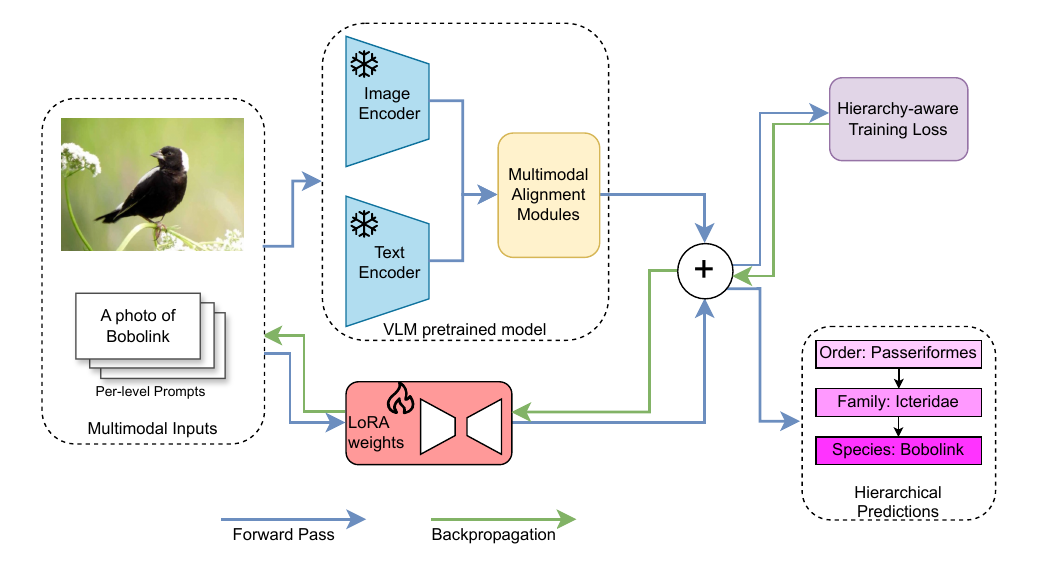}
    \caption{Overview of the proposed hierarchy-aware fine-tuning framework for Vision-Language Models.}
    \label{fig:overview}
\end{figure}

VLMs work well for captioning and detection but remain underexplored in hierarchical classification, where labels follow multi-level taxonomies (e.g., order → family → species). Most methods assume a flat label space, ignoring structural dependencies among categories. This produces inconsistent predictions across levels, limiting reliability in fine-grained domains like biodiversity monitoring and medical imaging.

We propose an efficient hierarchy-aware fine-tuning framework for VLMs using Tree-Path KL Divergence (TP-KL) and Hierarchy-Sibling Smoothed Cross-Entropy (HiSCE) losses. TP-KL enforces vertical consistency by matching the predicted label path through the taxonomy with ground truth, while HiSCE encourages horizontal consistency by smoothing probability distributions among sibling classes. Both losses work in the shared image–text embedding space, integrating seamlessly with lightweight EFT methods like LoRA without modifying the pretrained architecture.

Experiments across four hierarchical benchmarks (CUB-200-2011~\cite{CUBwang2023consistency_200_2011}, FGVC-Aircraft ~\cite{maji2013finegrainedvisualclassificationaircraft}, Butterfly-200~\cite{chen2018fine}, ChestX-ray14~\cite{wang2017chestx}) show that our approach improves Full Path Accuracy (FPA) and reduces Tree-based Inconsistency Error (TICE) compared to standard cross-entropy fine-tuning. Qualitative analysis shows that joint TP-KL + HiSCE optimization produces hierarchy-aligned semantic clusters, improving structural awareness in VLMs.

Our main contributions are:

\begin{itemize}
    \item The first efficient fine-tuning framework that explicitly models hierarchical label dependencies in VLMs.

    \item Dual-consistency objectives (TP-KL and HiSCE) that jointly enforce vertical and horizontal consistency across taxonomy levels.

    \item Comprehensive evaluation across multiple VLM architectures and datasets showing consistent improvements in hierarchy-aware metrics with minimal parameter overhead.
\end{itemize}

%% file: content/related_work.tex
\section{Related Work}

\subsection{Hierarchical Classification}
Hierarchical classification investigates visual categorization when labels follow multi-level taxonomies, such as order → family → species in biodiversity datasets or coarse-to-fine disease groupings in medical imaging ~\cite{bjerge2023hierarchical,elhamod2022hierarchy,yu2025hierarchical,sali2020hierarchical}. Classical methods ~\cite{zhu2017b,li2021mmf,valmadre2022hierarchical} model the hierarchy explicitly using multi-branch architectures, hierarchical softmax, or structured prediction modules. Recent approaches incorporate hierarchy-awareness through embedding structures, attention mechanisms, and graph-based modeling. ~\cite{chen2018fine} introduces coarse-to-fine feature refinement using taxonomy-guided supervision, while ~\cite{wang2023consistency} enforces vertical path coherence to reduce contradictory predictions across levels. Attention-based models ~\cite{liu2022focus} capture cross-level dependencies to improve fine-grained discrimination. Graph-based formulations ~\cite{elhamod2022hierarchy} including hierarchical graph convolution networks and taxonomy-aware GNNs, which propagate semantic information along parent–child edges to enrich label-aware representations. 
More recent work, such as visually consistent hierarchical image classification ~\cite{park2025visuallyconsistenthierarchicalimage}, aligns feature space geometry with taxonomy structure using contrastive constraints.


However, nearly all previous work relies on unimodal backbones, including EfficientNet-V2~\cite{tan2021efficientnetv2smallermodelsfaster} and DeiT-Tiny ~\cite{pmlr-v139-touvron21a}, which achieve strong fine-grained accuracy but lack cross-modal priors from large-scale image–text pretraining. In addition, most methods treat levels independently, which often leads to inconsistent predictions. Our work differs by enforcing both vertical (path-level) and horizontal (sibling-level) consistency directly within a multimodal embedding space of a VLM using customized training loss and smooth labeling.

\subsection{Efficient Fine-Tuning of Vision-Language Models}
Vision-Language Models (VLMs) leverage large-scale contrastive learning to align images and text in a shared embedding space~\cite{li2022blip, radford2021learning, yu2022coca, li2021align,yang2022vision}. 
While VLMs generalize well in zero-shot settings, adapting them to domain-specific tasks is computationally expensive due to their large number of parameters~\cite{lester2021power}. Efficient Fine-tuning (EFT) methods, including Prompt Tuning~\cite{prompt, jin2022goodpromptworthmillions}, Adapter Tuning~\cite{adapter}, LoRA~\cite{lora}, and QLoRA~\cite{qlora}, address this challenge by updating only a small subset of parameters while keeping pretrained backbones frozen. Prompt Tuning learns a small set of continuous prompt vectors prepended to the text tokens, guiding the frozen model toward the downstream task with minimal overhead. Adapter Tuning inserts lightweight bottleneck layers inside transformer blocks, allowing task-specific transformations while preserving all pretrained weights. LoRA introduces low-rank decomposition into attention projections, providing memory-efficient weight updates without modifying the original parameters. QLoRA further improves efficiency by quantizing the pretrained backbone to 4-bit while applying LoRA updates on top.



\subsection{Hierarchy-Aware Learning in Embedding-Space Models}

Taxonomy-aware learning can be achieved through loss-based regularization, which introduces structural priors directly into the optimization objective. Works such as hierarchical attention modeling~\cite{liu2022focus, hu2025multi} and hierarchical semantic embedding~\cite{chen2018fine,mumtaz2022hierarchy} explore ways to incorporate relationships among parent–child and sibling categories. However, these approaches operate primarily on unimodal embeddings and do not leverage multimodal representations.

The recently proposed Tree-Path KL Divergence ~\cite{park2025visuallyconsistenthierarchicalimage} enforces global vertical consistency by aligning predicted probability paths with ground-truth label paths, while smoothing-based methods distribute probability mass among semantically related sibling classes. In our work, we extend these ideas into a multimodal embedding space.


%% file: content/method.tex
\section{Methodology}
\textbf{Problem Statement}: Given an input image $x$ and its hierarchical labels $\{y^{(1)}, y^{(2)}, \ldots, y^{(L)}\}$ across $L$ taxonomy levels, where $y^{(1)}$
 denotes the coarsest level with the fewest categories (e.g. order, family, supercategory) and $y^{(L)}$
 denotes the finest level with the most categories (e.g. species, variant, specific disease),
our goal is to adapt a pretrained vision–language model to produce hierarchy-consistent predictions under efficient fine-tuning constraints. 


To explicitly incorporate taxonomy structure into training, we introduce combining two complementary loss functions:
(1) \textbf{Hierarchy-Sibling Smoothed Cross-Entropy (HiSCE)} encourages \emph{horizontal} consistency by smoothing the label distribution among sibling classes within the same level, improving robustness against fine-grained misclassifications; and 
(2) \textbf{Tree-Path KL Divergence (TP-KL)} enforces \emph{vertical} consistency by aligning the predicted path across hierarchy levels with the ground-truth taxonomy. 
The overall training objective combines these components with the standard cross-entropy loss:
\[
\mathcal{L}_{total} = \mathcal{L}_{CE} + \lambda_{1}\mathcal{L}_{TP\text{-}KL} + \lambda_{2}\mathcal{L}_{HiSCE},
\]
where $\lambda_{1}$ and $\lambda_{2}$ are weights of training loss for TP-KL and HiSCE respectively. 
The following subsections introduce the details of each component. We later combine TP-KL with HiSCE in Section~\ref{sec:joint} to demonstrate their complementarity.

\subsection{Hierarchy-Sibling Smoothed Cross-Entropy Loss (HiSCE)}
To enhance intra-level consistency, we apply a loss function called \textbf{Hierarchy-sibling Smoothed Cross-Entropy Loss (HiSCE)}. Traditional cross-entropy loss assigns a probability of 1 to the ground-truth label and 0 to all others, which overlooks the semantic similarity among classes under the same parent node. In contrast, our method introduces a hierarchy-aware smoothing mechanism that distributes a small portion of probability to the sibling (or cousin) classes that share the same parent, effectively leveraging structural similarity within each level. We construct a per-level label-smoothing table $T^{(l)}$, where each row corresponds to the smoothed target distribution of one class. Although this table can be viewed conceptually as a list of $C_l$ one-dimensional target vectors, representing each class’s softened label distribution, we implement it as a square matrix $T^{(l)} \in \mathbb{R}^{C_l \times C_l}$ for efficient batched indexing in modern deep-learning frameworks.

Formally, the $i$-th row of $T^{(l)}$ redistributes a small probability mass $\varepsilon_l$ uniformly among the siblings of class $i$, while assigning $1-\varepsilon_l$ to the diagonal entry as the ground truth. During training, given the ground-truth class index, we retrieve the entire smoothed target vector simply by selecting the corresponding row of $T^{(l)}$. 
 

\begin{equation}
    T^{(l)}_{ij} = 
    \begin{cases}
        1 - \varepsilon_l, &  \text{if } i=j, \\[4pt]
        \frac{\varepsilon_l}{|\mathcal{S}(i)|}, & \text{if } j \in \mathcal{S}(i),\\[4pt]
        0, & \text{otherwise,}
       
    \end{cases}
\end{equation}

where $\mathcal{S}(i)$ denotes the set of sibling classes of class $i$. 

\begin{figure}[htbp]
    \centering
\begin{forest}
for tree={
    align=center,
    edge={-, line width=0.9pt},
    l sep=12pt,
    s sep=10pt,
    parent anchor=south,
    child anchor=north,
    anchor=center,
}
[R
  [Parent1
    [\underline{\textbf{B}}\\{\small $= 1 - \epsilon$}]
    [C\\{\small $= \epsilon/2$}]
    [D\\{\small $= \epsilon/2$}]
  ]
  [Parent2
    [F\\{\small = 0}]
    [G\\{\small = 0}]
  ]
]
\end{forest}
\caption{Illustration of hierarchy-sibling smoothing at a single taxonomy level.
    The ground-truth class \underline{\textbf{B}} retains probability
    $1-\epsilon$, its sibling classes $C$ and $D$ share the remaining mass
    $\epsilon$ uniformly, and all non-sibling classes ($F$, $G$) receive $0$.
    This corresponds to one row of the smoothing matrix $T^{(l)}$ in 
    Equation~(1).}
\label{fig:HiSCE}
\end{figure}
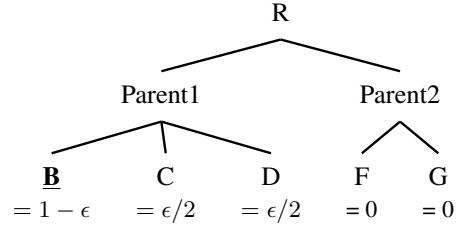

As shown in Figure~\ref{fig:HiSCE}, HiSCE encourages the model to learn \textit{smooth decision boundaries} within each level of the taxonomy by softly rewarding predictions close to the correct label in semantic space. As a result, the model can better capture visual and contextual similarities between related categories, improving robustness to fine-grained misclassifications and enhancing hierarchy-consistent generalization.

\subsection{Tree-path KL Divergence Loss (TP-KL)}

\begin{figure*}[t]
    \centering
    \includegraphics[width=1\linewidth]{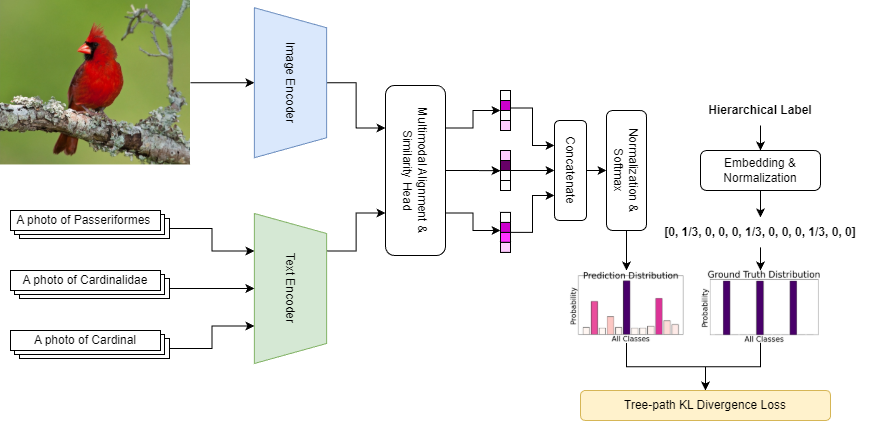}
    \caption{Overview of the Tree-Path KL Divergence computation.}
    \label{fig:tp_kl}
\end{figure*}

\textbf{Tree-Path KL Divergence (TP-KL)} loss was originally proposed as a model-agnostic approach to enforce \emph{semantic consistency across hierarchy levels}. The key idea is to align the predicted hierarchical path of a sample with its ground-truth label path through the taxonomy. In conventional classification architectures (e.g., CNNs or Transformers), each hierarchy level is equipped with an independent classification head producing logits over classes. TP-KL first concatenates the log-softmax outputs from all levels to form a joint hierarchical distribution and then computes the Kullback–Leibler (KL) divergence between this predicted distribution and the concatenated one-hot ground-truth path.

Formally, given $L$ hierarchy levels, each with logits $z^{(l)}$ and one-hot label vector $\mathbf{1}_{y^{(l)}}$, 
the TP-KL loss is defined as:
\begin{equation}
\mathrm{Y} = \tfrac{1}{L}[\mathbf{1}_{y^{(L)}}; \ldots; \mathbf{1}_{y^{(1)}}]
\end{equation}


\begin{equation}
    \mathcal{L}_{\text{TP-KL}} = \mathrm{KL}\!\left(
\!\mathrm{Softmax}\big([\mathbf{z}^{(L)}; \ldots; \mathbf{z}^{(1)}]\big), \, Y
\right)
\end{equation}

This loss penalizes predictions that violate the hierarchical structure by encouraging consistent outputs along the tree path from root to leaf nodes.

When applying TP-KL to VLM, the loss formulation remains unchanged; the adaptation lies only in how the prediction distributions are obtained. Instead of using classifier logits from independent heads, VLMs represent categories as \emph{text embeddings} and compute \emph{image–text similarity scores} in a shared embedding space. For each hierarchy level $l$, the similarity for class $c$ is computed as
\begin{equation}
\mathbf{z}^{(l)}_c = \frac{\mathbf{v}^\top \mathbf{t}^{(l)}_c}{\|v\|\|\mathbf{t}^{(l)}_c\|}
\end{equation}

where $v$ and $t^{(l)}_c$ denote normalized image and text embeddings, respectively. The similarity scores of each level are normalized with a temperature-scaled logarithmic softmax~\cite{hinton2015} and concatenated to form the hierarchical distribution used in the KL divergence. This alignment allows TP-KL to operate directly in the embedding space, making it naturally compatible with EFT strategies such as LoRA or adapter tuning, without altering the original loss formulation.

Figure~\ref{fig:tp_kl} visualizes how hierarchical predictions from different taxonomy levels are concatenated to form a unified probability path before computing the KL divergence against the ground-truth path distribution. It shows that predictions from all hierarchy levels, e.g., order → family → species, are first projected into the shared embedding space and then aligned through temperature-scaled log-softmax operations. The resulting joint distribution reflects the complete hierarchical path, enforcing vertical consistency during training.

%% file: content/experiment.tex
\section{Experiments}
We evaluate our hierarchy-aware fine-tuning framework across four datasets with three different model architectures. We first introduce the datasets, then the evaluation metrics, and finally analyze each component.

\subsection{Dataset Overview}
Table~\ref{tab:dataset} summarizes the hierarchical datasets used for evaluation.
These datasets span biodiversity, aviation, and medical imaging, with hierarchical label structures of two to four taxonomy levels.

\begin{table*}[!htbp]
    \centering
    \begin{tabularx}{\textwidth}{lclcc}
    \hline\hline
        \textbf{Dataset} & \textbf{\#Class} & \textbf{Hierarchy Label (High $\rightarrow$ Low)} & \textbf{Levels} & \textbf{Domain}
        \\ \hline
        CUB-200-2011~\cite{CUBwang2023consistency_200_2011} & 200 & order: 13, family: 38, species: 200 & 3 & Biodiversity \\
        FGVC-Aircraft~\cite{maji2013finegrainedvisualclassificationaircraft} & 100 & manufacturer: 30, family: 70, variant: 100 & 3 & Aircraft \\
        ChestX-ray14~\cite{wang2017chestx} & 14 & supercategory: 3, subcategory: 9, specific disease: 14 & 3 & Medical \\
        Butterfly-200~\cite{chen2018fine} & 200 & family: 5, subfamily: 23, genus: 116, species: 200 & 4 & Biodiversity \\
    \hline\hline
    \end{tabularx}
    \caption{Statistics of hierarchical classification datasets used in our experiments.
    Each dataset provides multi-level taxonomies ranging from coarse to fine categories.}
    \label{tab:dataset}
\end{table*}

\noindent\textbf{CUB-200-2011}~\cite{CUBwang2023consistency_200_2011} is a bird-species benchmark with
11{,}788 images (5{,}994 training / 5{,}794 test). Since the original annotations only provide species-level information, following ~\cite{chang2021your}, we add two hierarchy levels (family and order) to create a 3-level label chain: order → family → species.
\textbf{FGVC-Aircraft}~\cite{maji2013finegrainedvisualclassificationaircraft} contains 10{,}000 aircraft images at three levels: manufacturer, family, and variant.
\textbf{ChestX-ray14}~\cite{wang2017chestx} includes over 112{,}000 chest X-ray images from 30{,}000 patients in a three-level disease taxonomy.
\textbf{Butterfly-200}~\cite{chen2018fine} has $\sim$25{,}000 butterfly images covering 200 species in a four-level hierarchy (family $\rightarrow$ subfamily $\rightarrow$ genus $\rightarrow$ species).

\subsection{Evaluation Metrics}
We evaluate using accuracy-based metrics and hierarchy-aware consistency metrics.

\textbf{Accuracy} measures the proportion of correctly classified instances at each level, macro-averaged over all levels.

\textbf{Weighted Average Precision (wAP)~\cite{liu2022focus}} assigns greater weight to accuracy at fine-grained levels:
\begin{equation}
\mathrm{wAP} = \sum_{l=1}^{L} \frac{N_l}{\sum_{k=1}^{L} N_k} P_l,
\end{equation}
where $N_l$ and $P_l$ denote the number of classes and precision at level $l$. wAP emphasizes performance on deeper, fine-grained levels.

\textbf{Tree-based InConsistency Error rate (TICE)~\cite{wang2023consistency}} measures whether the prediction path is valid in the hierarchical tree. $\mathrm{TICE} = n_{ic} / N$, where $n_{ic}$ is the number of inconsistent prediction paths and $N$ is the total number of predictions.

\textbf{Full Path Accuracy(FPA)~\cite{park2025visuallyconsistenthierarchicalimage}} evaluates overall accuracy and hierarchical consistency. $\mathrm{FPA} = n_{ac} / N$, where $n_{ac}$ is the number of samples correct across all hierarchical levels. FPA measures the proportion of instances with entirely correct label paths.

\subsection{Baseline Configuration: LoRA-based VLM}\label{sec:lora}
We adopt Low-Rank Adaptation (LoRA) ~\cite{lora} as our efficient fine-tuning strategy on CLIP ~\cite{transferablevisualmodels}.

LoRA introduces lightweight rank-decomposition matrices into linear projection layers of both vision and text transformers.
Each weight matrix $W \in \mathbb{R}^{d \times k}$ gets a low-rank update $\Delta W = B A$, where $ A \in \mathbb{R}^{r \times k}, \; B \in \mathbb{R}^{d \times r}$. The rank $r$ controls capacity (lower $r$ means fewer trainable parameters), while scaling factor $\alpha$ balances the update magnitude relative to frozen base weights. This reduces GPU memory and fine-tuning cost while maintaining generalization benefits.

We set rank $r=16$, scaling factor $\alpha=32$, and dropout $p=0.3$. Base model weights stay frozen; only LoRA parameters and layer normalization weights are trainable (~4.4 million parameters). We use AdamW optimizer ~\cite{adamw} with learning rate 1e-3 and train for 100 epochs (Table~\ref{tab:env}).

\begin{table}[!htbp]
\centering
\caption{Experimental Environment Setup}
\begin{tabular}{ll}
\hline
\textbf{Component} & \textbf{Specification} \\
\hline
GPU & NVIDIA RTX 6000  \\
CUDA / Driver & CUDA 13.0 / Driver 581.42 \\
CPU & Intel Xeon Silver 4316 (2.3\,GHz) \\
Memory & 128\,GB RAM \\
Operating System & Ubuntu 22.04 LTS \\
Software & PyTorch 2.3, OTX ~\cite{openvino_otx}\\
\hline
\end{tabular}
\label{tab:env}
\end{table}

Table~\ref{tab:clip_lora} shows that CLIP-LoRA achieves strong hierarchical classification performance despite few trainable parameters. Compared to fully fine-tuned backbones like EfficientNet-V2 and DeiT-Tiny, CLIP-LoRA shows competitive or superior Full-Path Accuracy (FPA) and hierarchical consistency (TICE) while maintaining high efficiency. These results validate that LoRA tuning preserves VLM generalization and provides a reliable foundation for hierarchy-aware fine-tuning.

\begin{table*}[!htbp]
\centering
\caption{Performance comparison between different configurations.}
\begin{tabular}{lll|rrlr}
\hline\hline

Architecture & Trainable Params & Dataset & Accuracy & FPA & TICE $\downarrow$ & wAP \\ \hline
\multirow{4}{*}{EfficientNet\_v2~\cite{tan2021efficientnetv2smallermodelsfaster}} & \multirow{4}{*}{20.2M} &
FGVC-Aircraft ~\cite{maji2013finegrainedvisualclassificationaircraft} & \textbf{71.7} & \textbf{55.7} & 20.4 & \textbf{71.2} \\
&& ChestX-ray14 ~\cite{wang2017chestx} & 59.7 & 31.7 & -- & 26.6 \\
&& CUB-200-2011 ~\cite{CUBwang2023consistency_200_2011} & \textbf{84.3} & \textbf{71.0} & -- &\textbf{ 82.5} \\
&& Butterfly-200 ~\cite{chen2018fine}  & \textbf{88.4} & \textbf{73.9} & -- & \textbf{83.6} \\ \hline

\multirow{4}{*}{deit\_tiny~\cite{pmlr-v139-touvron21a}} & \multirow{4}{*}{5.6M} &
FGVC-Aircraft ~\cite{maji2013finegrainedvisualclassificationaircraft} & 42.0 & 21.3 & 54.9 & 41.0 \\
 &  & ChestX-ray14 ~\cite{wang2017chestx} & 60.5 & 31.8 & -- & \textbf{28.1} \\
 &  & CUB-200-2011 ~\cite{CUBwang2023consistency_200_2011} & 61.2 & 40.0 & -- & 54.8 \\
 &  & Butterfly-200 ~\cite{chen2018fine} & 76.3 & 54.3 & -- & 65.1 \\ \hline

\multirow{4}{*}{CLIP-LoRA~\cite{radford2021learning}} & \multirow{4}{*}{4.4M} &
FGVC-Aircraft ~\cite{maji2013finegrainedvisualclassificationaircraft} & 53.0 & 38.3 & \textbf{17.9} & 55.0 \\
 &  & ChestX-ray14 ~\cite{wang2017chestx} &\textbf{ 65.8} & \textbf{36.5} & \textbf{11.8} & 22.9 \\
 &  & CUB-200-2011 ~\cite{CUBwang2023consistency_200_2011} & 71.7 & 50.2 & 21.9 & 68.8 \\
 &  & Butterfly-200 ~\cite{chen2018fine} & 83.4 & 72.1 & 10.1 & 79.2 \\

\hline\hline
\end{tabular}
\label{tab:clip_lora}
\end{table*}

\subsection{Tree-path KL Divergence Loss}

We evaluate Tree-Path KL Divergence (TP-KL) loss on the CLIP-LoRA model (Section~\ref{sec:lora}). TP-KL combines with cross-entropy (CE) as
\begin{equation}
    L = L_{CE} + \lambda L_{TP-KL}
\end{equation}
where $\lambda$ controls hierarchical regularization. We test $\lambda \in \{0, 0.5, 1,2, 5\}$ across four datasets: CUB-200-2011, FGVC-Aircraft, ChestX-ray14, and Butterfly-200.

\begin{figure}[!h]
    \centering
    \begin{subfigure}[b]{0.45\textwidth}
        \includegraphics[width=\textwidth]{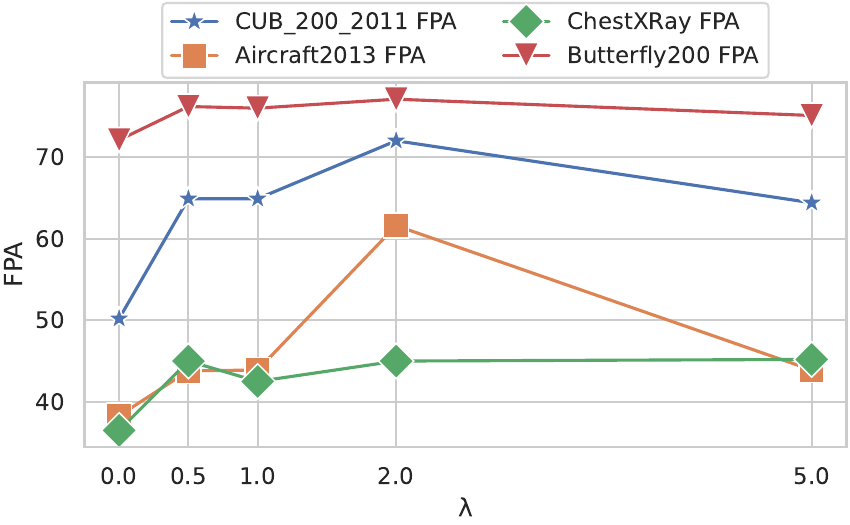}
    \end{subfigure}
    \hfill

    \begin{subfigure}[b]{0.45\textwidth}
        \includegraphics[width=\textwidth]{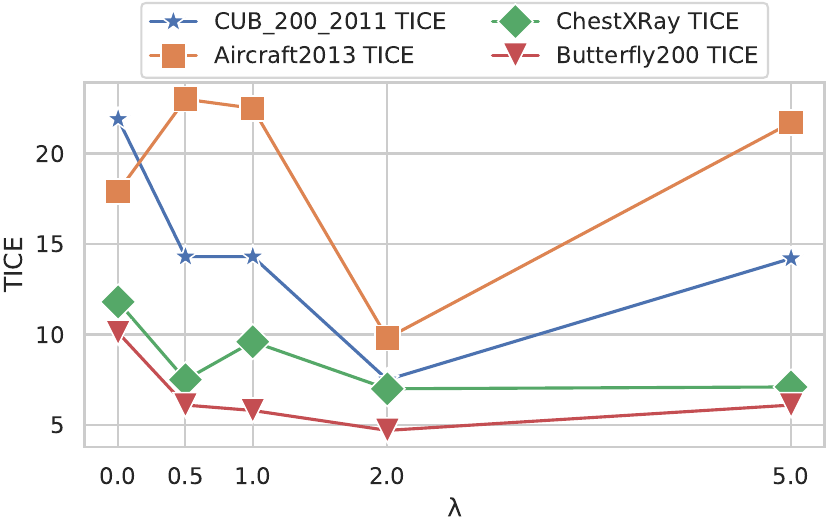}
    \end{subfigure}

    \caption{Performance comparison across datasets under different TP-KL weights $\lambda$: FPA (top) and TICE (bottom).}
    \label{fig:kl_weight}
\end{figure}

Table~\ref{tab:kl_weight} shows that TP-KL substantially improves Full-Path Accuracy (FPA) and reduces Tree-based Inconsistency Error (TICE) across all domains. Best results use moderate regularization ($\lambda = 2$), with +21.8 FPA and –14.4 TICE on CUB-200-2011, and similar improvements on Butterfly-200 and FGVC-Aircraft. Even on large-scale ChestX-ray14, TP-KL enhances hierarchical consistency while preserving accuracy. Figure~\ref{fig:kl_weight} shows results across datasets for different $\lambda$ values.

\begin{table}[!htbp]
    \centering
    \caption{Effect of Tree-Path KL Divergence Loss Weights on CLIP-LoRA}
    \label{tab:kl_weight}
    \resizebox{\linewidth}{!}{%
    \begin{tabular}{llccccc}
        \toprule
        Dataset & Metric & \multicolumn{5}{c}{CE + \(\lambda\) * KL} \\
        \cmidrule(lr){3-7}
        & & \(\lambda = 0\) & \(\lambda = 0.5\) & \(\lambda = 1\) & \(\lambda = 2\) & \(\lambda = 5\) \\
        \midrule
        \multirow{5}{*}{\textbf{CUB-200-2011}~\cite{CUBwang2023consistency_200_2011}}
        &Accuracy & 71.7 & 81.8 & 81.7 &\textbf{ 83.6} & 81.3 \\
        &FPA & 50.2 & 64.9 & 64.9 & \textbf{72} & 64.4 \\
        &TICE$\downarrow$ & 21.9 & 14.3 & 14.3 & \textbf{7.5}& 14.2 \\
        &wAP & 68.8 & 75.7 & 75.8 & \textbf{80.5}& 75.6 \\
        \midrule
        \multirow{5}{*}{\textbf{FGVC-Aircraft}~\cite{maji2013finegrainedvisualclassificationaircraft}}
        &Accuracy & 53 & 58 & 58.1& \textbf{68.6}& 58.1 \\
        &FPA & 38.3 & 43.8 & 43.9 &\textbf{61.6}& 43.9 \\
        &TICE$\downarrow$ & 17.9 & 23 & 22.5 &\textbf{9.8}& 21.7 \\
        &wAP & 55 & 59.1 & 59.3 &\textbf{72.6}& 59.9 \\
        \midrule
        \multirow{5}{*}{\textbf{ChestX-ray14}~\cite{wang2017chestx}}
        &Accuracy &\textbf{ 65.8} & 65 & 66.1 &64.9& 64.9 \\
        &FPA & 36.5 & 45 & 42.5 &45& \textbf{45.2} \\
        &TICE$\downarrow$ & 11.8 & 7.6 & 7.6 &\textbf{7}& 7.9 \\
        &wAP & 22.9 & 21.5 & 22.5 &\textbf{25}& 23.8 \\
        \midrule
        \multirow{5}{*}{\textbf{Butterfly-200}~\cite{chen2018fine}}
        &Accuracy & 83.4 & 84.7 & 84.4 &\textbf{86.6}& 83.7 \\
        &FPA & 72.1 & 76.2 & 76 &\textbf{77.1}& 75.1 \\
        &TICE$\downarrow$ & 10.1 & 6.1 & 5.8 &\textbf{4.7}& 6.1 \\
        &wAP & 79.2 & 82.3 & 82.6 &\textbf{83.1}& 81.2 \\
        \bottomrule
    \end{tabular}}
\end{table}

We also evaluate TP-KL on a conventional vision backbone to show its generality.
We fine-tune EfficientNet\_v2 with different KL weights ($\lambda=0,1,5$) on CUB-200-2011 and FGVC-Aircraft. Table~\ref{tab:effnet_kl} shows that adding KL improves accuracy and hierarchical consistency, demonstrating that TP-KL benefits conventional vision models too.

\begin{table}[!htbp]
    \centering
    \caption{TP-KL on EfficientNet\_v2 with different loss weights.}
    \label{tab:effnet_kl}
    \resizebox{\linewidth}{!}{%
    \begin{tabular}{ll|ccc|ccc}
        \toprule
        & & \multicolumn{3}{c|}{CUB-200-2011} & \multicolumn{3}{c}{FGVC-Aircraft} \\
        & Metric & $\lambda{=}0$ & $\lambda{=}1$ & $\lambda{=}5$ & $\lambda{=}0$ & $\lambda{=}1$ & $\lambda{=}5$ \\
        \midrule
        \multirow{3}{*}{\rotatebox{90}{ENet-v2}}
        & Acc. & 0.853 & 0.850 & \textbf{0.858} & 0.709 & 0.742 & \textbf{0.764} \\
        & FPA & 0.735 & 0.724 & \textbf{0.764} & 0.551 & 0.593 & \textbf{0.629} \\
        & wAP & \textbf{0.836} & 0.832 & 0.829 & 0.705 & 0.733 & \textbf{0.746} \\
        \bottomrule
    \end{tabular}
    }
\end{table}

\subsection{Hierarchy-Sibling Smoothed Cross-Entropy Loss (HiSCE)}
\begin{figure*}[!htbp]
    \centering
    \begin{subfigure}[t]{0.48\linewidth}
        \centering
        \includegraphics[width=\linewidth]{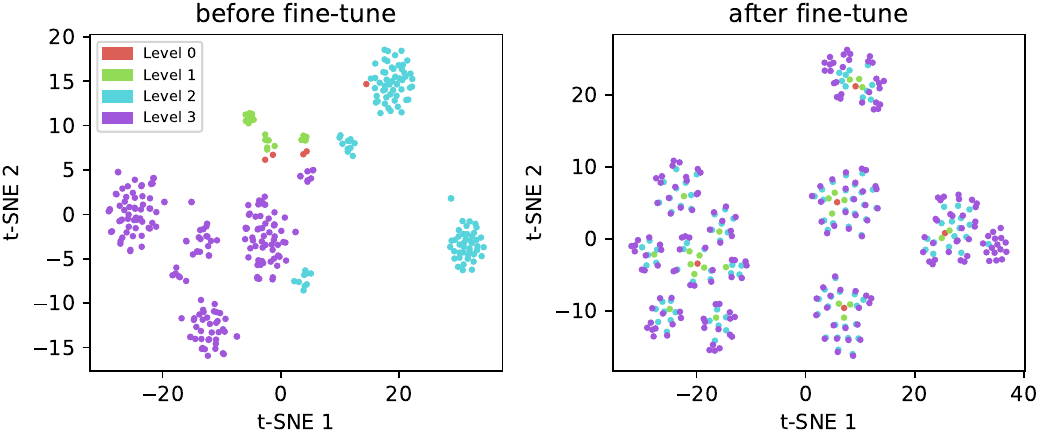}
        \caption{Butterfly-200 t-SNE visualization}
        \label{fig:tsne_hier}
    \end{subfigure}
    \hfill
    \begin{subfigure}[t]{0.48\linewidth}
        \centering
        \includegraphics[width=\linewidth]{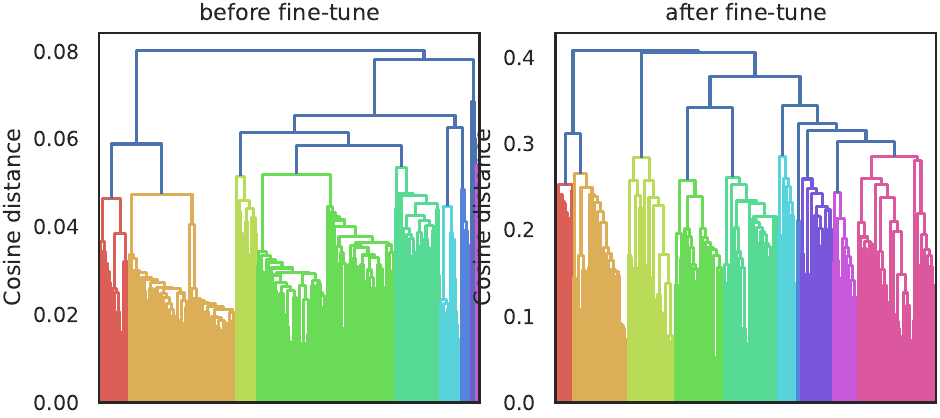}
        \caption{Butterfly-200 Dendrogram comparison}
        \label{fig:dendro_hier}
    \end{subfigure}

    \begin{subfigure}[t]{0.48\linewidth}
        \centering
        \includegraphics[width=\linewidth]{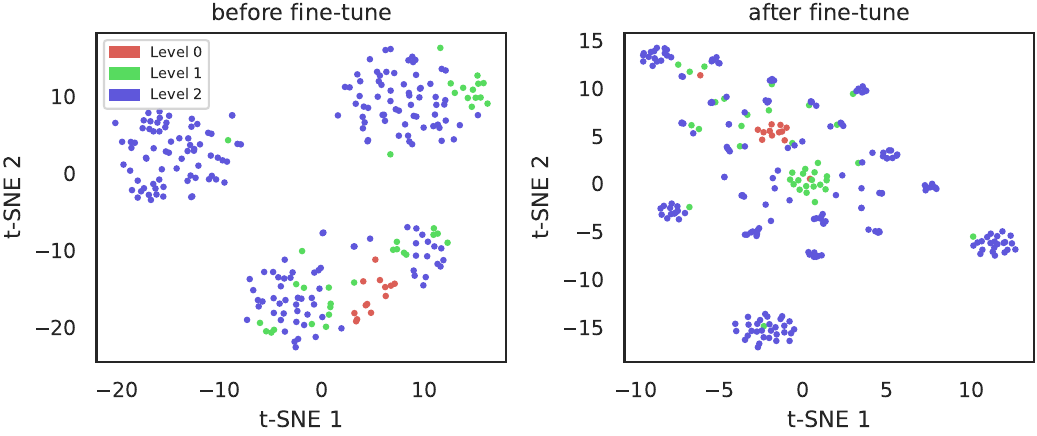}
        \caption{CUB-200-2011 t-SNE visualization}
        \label{fig:tsne_cub}
    \end{subfigure}
    \hfill
    \begin{subfigure}[t]{0.48\linewidth}
        \centering
        \includegraphics[width=\linewidth]{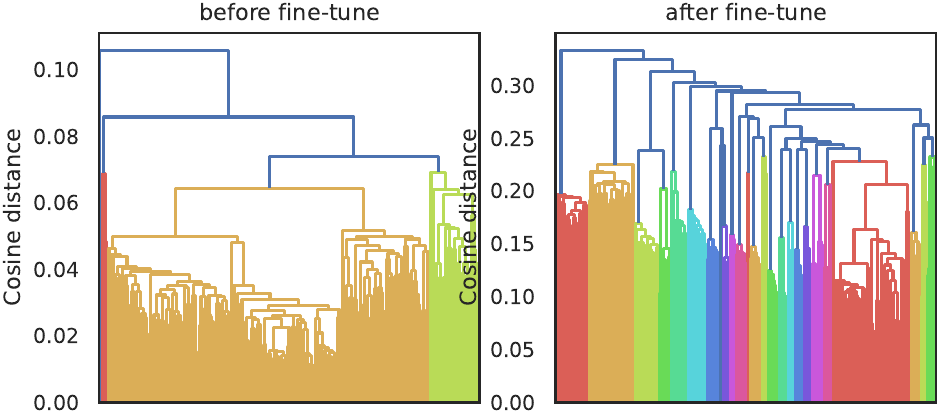}
        \caption{CUB-200-2011 Dendrogram comparison}
        \label{fig:dendro_cub}
    \end{subfigure}

    \vspace{-1ex}
    \caption{Visualization of text embeddings before and after joint TP-KL + HiSCE optimization on 2 datasets (Butterfly-200 and CUB-200-2011).
Left: t-SNE projections, where each point denotes a label embedding colored by its hierarchy level. Right: dendrograms visualizing hierarchical grouping, where branch colors indicate clustering structure.}
    \label{fig:hier_semantics}
\end{figure*}
We evaluate Hierarchy-Sibling Smoothed Cross-Entropy (HiSCE) on four datasets using CLIP-LoRA (Section~\ref{sec:lora}).
We compare four configurations:

\begin{enumerate}
    \item Zero-shot: pretrained CLIP without fine-tuning;
    \item CE: fine-tuning with standard cross-entropy;
    \item HiSCE: replacing CE with hierarchy-sibling smoothing;
    \item CE + HiSCE: combining both with equal weight (1.0).
\end{enumerate}

Table~\ref{tab:HiSCE} shows that HiSCE consistently enhances Full-Path Accuracy (FPA) and reduces Tree-based Inconsistency Error (TICE) across all datasets. CE + HiSCE achieves best performance, with notable improvements on fine-grained datasets like CUB-200-2011 (FPA: 50.2 → 71.1, TICE: 21.9 → 6.3). Similar trends appear in Butterfly-200, showing that sibling-level smoothing reinforces semantic coherence within each hierarchy level.

HiSCE complements Tree-Path KL Divergence (TP-KL) by addressing horizontal consistency across sibling categories, while TP-KL enforces vertical path consistency across levels. Together, these objectives provide a comprehensive, lightweight fine-tuning framework for adapting VLMs to structured taxonomies.
\begin{table}[!h]
    \centering
    \caption{Effect of Hierarchy-Sibling Smoothed Cross-Entropy on CLIP-LoRA}
    \label{tab:HiSCE}
    \resizebox{\linewidth}{!}{%
    \begin{tabular}{l|c|c|ccc}
        \toprule
       Dataset & & Zero-shot & CE & HiSCE & CE+HiSCE \\
        \midrule
        \multirow{4}{*}{\textbf{CUB-200-2011}~\cite{CUBwang2023consistency_200_2011}}
        &Accuracy & 23.4 & 71.7 & 81.8 & \textbf{84.1} \\
        &FPA & 0.5 & 50.2 & 63.1 &\textbf{ 71.1} \\
        &TICE$\downarrow$ & 98.3 & 21.9 & 10.8 & \textbf{6.3} \\
        &wAP & 44.9 & 68.8 & 72.5 &\textbf{ 79.1} \\
        \midrule
        \multirow{4}{*}{\textbf{FGVC-Aircraft}~\cite{maji2013finegrainedvisualclassificationaircraft}}
        &Accuracy & 19.8 & 53 & 49.5 & \textbf{65.1} \\
        &FPA & 4 & 38.3 & 32.7 & \textbf{57} \\
        &TICE{$\downarrow$} & 85.7 & 17.9 & 37.5 & \textbf{11.7} \\
        &wAP & 22.6 & 55 & 51.7 & \textbf{70.2} \\
        \midrule
        \multirow{4}{*}{\textbf{ChestX-ray14}~\cite{wang2017chestx}}
        &Accuracy & 5.3 & \textbf{65.8} & 65.5 & 65.6 \\
        &FPA & 0 & 36.5 & 39.7 & \textbf{43.8} \\
        &TICE{$\downarrow$} & 99.9 & 11.8 & 9.1 & \textbf{8.4} \\
        &wAP & 3.4 & 22.9 & 21.7 & \textbf{23.2} \\
        \midrule
        \multirow{4}{*}{\textbf{Butterfly-200}~\cite{chen2018fine}}
        &Accuracy & 13.5 & 83.4 & \textbf{85} & 84.7 \\
        &FPA & 0.8 & 72.1 & \textbf{77.4} & 76.5 \\
        &TICE{$\downarrow$} & 84 & 10.1 &\textbf{ 3.8} & 4.9 \\
        &wAP & 7.9 & 79.2 & \textbf{82.9} & 82.2 \\
        \bottomrule
    \end{tabular}}
\end{table}


\subsection{Joint TP-KL and HiSCE Optimization}\label{sec:joint}

TP-KL enforces vertical consistency along label hierarchies, while HiSCE promotes horizontal consistency within each level. TP-KL aligns predictions across levels but can over-penalize fine-grained confusion, whereas HiSCE smooths intra-level decisions but lacks global path alignment. We jointly optimize them with base cross-entropy to exploit their complementarity.

We use Optuna ~\cite{optuna} (100 trials) to find optimal weights $\lambda_1$ (TP-KL) and $\lambda_2$ (HiSCE) for each dataset, maximizing validation accuracy.
Table~\ref{tab:joint} shows ablation studies comparing individual losses with joint optimization.
TP-KL alone causes severe performance drops across datasets, showing that vertical regularization without intra-level smoothing over-penalizes fine-grained errors. HiSCE alone provides strong intra-level consistency but lacks global hierarchical alignment.
Joint optimization achieves best results, confirming that vertical (TP-KL) and horizontal (HiSCE) regularization complement each other, yielding hierarchy-aware representations that are both consistent and discriminative.

\begin{figure*}[!htbp]
  \centering

  \begin{subfigure}[b]{0.49\linewidth}
    \centering
    \includegraphics[width=\linewidth]{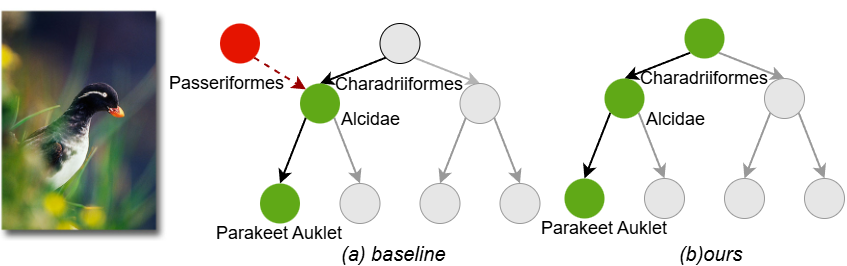}
    \label{fig:sub-a}
  \end{subfigure}
  \hfill
  \begin{subfigure}[b]{0.49\linewidth}
    \centering
    \includegraphics[width=\linewidth]{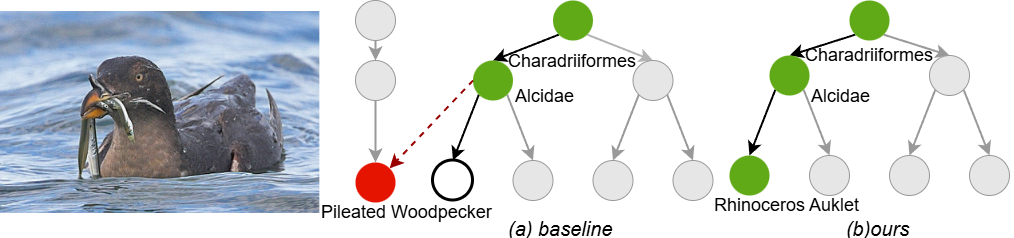}
    \label{fig:sub-b}
  \end{subfigure}

  \vspace{0.6em}

  \begin{subfigure}[b]{0.49\linewidth}
    \centering
    \includegraphics[width=\linewidth]{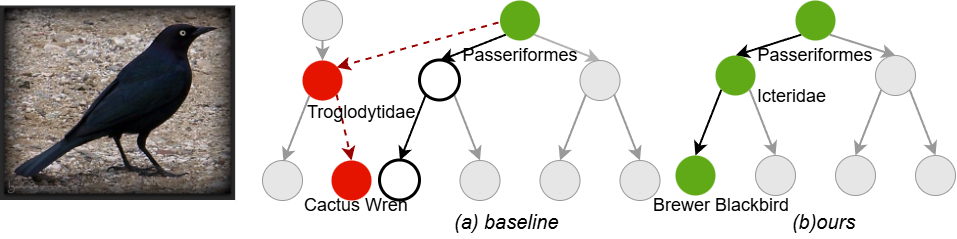}
    \label{fig:sub-c}
  \end{subfigure}
  \hfill
  \begin{subfigure}[b]{0.49\linewidth}
    \centering
    \includegraphics[width=\linewidth]{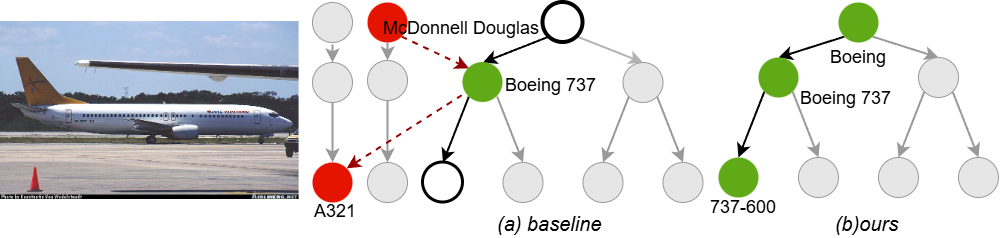}
    \label{fig:sub-d}
  \end{subfigure}

  \caption{Qualitative comparison of hierarchical predictions between the CLIP-LoRA baseline and our method.
Our approach corrects invalid or inconsistent label paths by enforcing vertical and sibling-level consistency across taxonomy levels.}
  \label{fig:compare}
\end{figure*}

\begin{table}[!htbp]
\centering
\caption{Performance comparison between different configurations.}
\resizebox{\linewidth}{!}{%
\begin{tabular}{cc|cccc}
\hline\hline

Dataset & Method & Acc. & FPA & TICE$\downarrow$ & wAP \\ \hline
\multirow{4}{*}{CUB-200-2011}
& TP-KL Only & 32.1&4.6&29.3&7.9 \\
& HiSCE only &81.8&63.1&10.8&72.5 \\
& Joint &\textbf{85.1}&\textbf{72.9}&\textbf{5.9}&\textbf{80.6}\\ \hline

\multirow{4}{*}{FGVC-Aircraft}
& TP-KL Only & 12.8&5.6&35.8&10.4 \\
& HiSCE only &49.5&32.7&37.5&51.7 \\
& Joint &\textbf{69.4}&\textbf{61.5}&\textbf{8.5}&\textbf{73.3}\\ \hline

\multirow{4}{*}{Butterfly-200}
& TP-KL Only & 32.5&13.5&36.7&17.0 \\
& HiSCE only &85&77.4&3.8&82.9 \\
& Joint &\textbf{87.1}&\textbf{77.5}&\textbf{3.2}&\textbf{82.9}\\ \hline

\multirow{4}{*}{ChestX-ray14}
& TP-KL Only & \textbf{67.8}&45.5&22.0&20.9 \\
& HiSCE only &65.5&39.7&\textbf{19.1}&\textbf{21.7} \\
& Joint &\textbf{67.8}&\textbf{46.2}&19.3&16.0\\

\hline\hline
\end{tabular}
}
\label{tab:joint}
\end{table}

Since all hierarchical levels in CLIP share the same vision encoder output, text embeddings carry hierarchy-specific semantics. Figure~\ref{fig:hier_semantics} visualizes text embeddings before and after fine-tuning using t-SNE projection ~\cite{JMLR:v9:vandermaaten08a} and hierarchical clustering (dendrogram). Before fine-tuning, label embeddings from different levels (e.g., order, family, species) are scattered and intermixed, showing that pretrained CLIP doesn't explicitly capture hierarchical dependencies. After joint TP-KL + HiSCE optimization, embeddings show clear multi-level grouping: fine-grained labels cluster around their coarse-level ancestors, and the structure aligns with ground-truth taxonomy. This confirms that joint objectives encourage the text encoder to internalize hierarchical relationships, transforming the textual embedding space into a semantically consistent manifold. Moreover, Figure \ref{fig:compare} illustrates representative prediction examples comparing the CLIP-LoRA baseline with our hierarchy-aware fine-tuning method. While the baseline frequently produces semantically inconsistent label paths that violate parent–child relationships, our approach yields predictions that are coherent across hierarchy levels. By jointly enforcing vertical path alignment and sibling-level consistency, TP-KL and HiSCE reduce error propagation from fine to coarse categories, resulting in more stable and interpretable hierarchical predictions.

%% file: content/conclusion.tex
\section{Conclusion}

We introduced an efficient fine-tuning framework for vision-language models on hierarchical classification. Two complementary objectives—Tree-Path KL Divergence (TP-KL) for vertical path consistency and Hierarchy-Sibling Smoothed Cross-Entropy (HiSCE) for horizontal intra-level coherence—enable VLMs to learn taxonomy-consistent representations while updating few parameters. Experiments across four domains show that our approach improves full-path accuracy and reduces hierarchical inconsistency compared to standard cross-entropy, while maintaining LoRA efficiency. Joint TP-KL + HiSCE optimization reshapes the text-embedding space into a hierarchy-aligned manifold, bridging coarse- and fine-grained semantics.

\textbf{Limitations.} Our approach assumes complete and balanced taxonomies, which may limit performance on noisy or sparse hierarchies. We evaluate only on classification tasks, leaving other multimodal settings unexplored.

\textbf{Future Work.} Future directions include handling dynamic or incomplete hierarchies, scaling to open-vocabulary settings, and extending hierarchy-aware objectives to generative or retrieval tasks.

%% file: ref.bib
@misc{prompt,
      title={Language Models are Few-Shot Learners},
      author={Tom B. Brown and Benjamin Mann and Nick Ryder and Melanie Subbiah and Jared Kaplan and Prafulla Dhariwal and Arvind Neelakantan and Pranav Shyam and Girish Sastry and Amanda Askell and Sandhini Agarwal and Ariel Herbert-Voss and Gretchen Krueger and Tom Henighan and Rewon Child and Aditya Ramesh and Daniel M. Ziegler and Jeffrey Wu and Clemens Winter and Christopher Hesse and Mark Chen and Eric Sigler and Mateusz Litwin and Scott Gray and Benjamin Chess and Jack Clark and Christopher Berner and Sam McCandlish and Alec Radford and Ilya Sutskever and Dario Amodei},
      year={2020},
      eprint={2005.14165},
      archivePrefix={arXiv},
      primaryClass={cs.CL},
      url={https://arxiv.org/abs/2005.14165},
}

@misc{adapter,
      title={Parameter-Efficient Transfer Learning for NLP},
      author={Neil Houlsby and Andrei Giurgiu and Stanislaw Jastrzebski and Bruna Morrone and Quentin de Laroussilhe and Andrea Gesmundo and Mona Attariyan and Sylvain Gelly},
      year={2019},
      eprint={1902.00751},
      archivePrefix={arXiv},
      primaryClass={cs.LG},
      url={https://arxiv.org/abs/1902.00751},
}

@misc{lora,
      title={LoRA: Low-Rank Adaptation of Large Language Models},
      author={Edward J. Hu and Yelong Shen and Phillip Wallis and Zeyuan Allen-Zhu and Yuanzhi Li and Shean Wang and Lu Wang and Weizhu Chen},
      year={2021},
      eprint={2106.09685},
      archivePrefix={arXiv},
      primaryClass={cs.CL},
      url={https://arxiv.org/abs/2106.09685},
}

@misc{transferablevisualmodels,
      title={Learning Transferable Visual Models From Natural Language Supervision},
      author={Alec Radford and Jong Wook Kim and Chris Hallacy and Aditya Ramesh and Gabriel Goh and Sandhini Agarwal and Girish Sastry and Amanda Askell and Pamela Mishkin and Jack Clark and Gretchen Krueger and Ilya Sutskever},
      year={2021},
      eprint={2103.00020},
      archivePrefix={arXiv},
      primaryClass={cs.CV},
      url={https://arxiv.org/abs/2103.00020},
}

@misc{jin2022goodpromptworthmillions,
      title={A Good Prompt Is Worth Millions of Parameters: Low-resource Prompt-based Learning for Vision-Language Models},
      author={Woojeong Jin and Yu Cheng and Yelong Shen and Weizhu Chen and Xiang Ren},
      year={2022},
      eprint={2110.08484},
      archivePrefix={arXiv},
      primaryClass={cs.CV},
      url={https://arxiv.org/abs/2110.08484},
}

@misc{qlora,
      title={QLoRA: Efficient Finetuning of Quantized LLMs},
      author={Tim Dettmers and Artidoro Pagnoni and Ari Holtzman and Luke Zettlemoyer},
      year={2023},
      eprint={2305.14314},
      archivePrefix={arXiv},
      primaryClass={cs.LG},
      url={https://arxiv.org/abs/2305.14314},
}

@techreport{CUBwang2023consistency_200_2011,
    title = {The Caltech-UCSD Birds-200-2011 Dataset},
    author = {Wah, C. and Branson, S. and Welinder, P. and Perona, P. and Belongie, S.},
    year = {2011},
    institution = {California Institute of Technology},
    number = {CNS-TR-2011-001}
}

@article{li2022blip,
  title={BLIP: Bootstrapping Language-Image Pre-training for Unified Vision-Language Understanding and Generation},
  author={Li, Junnan and Li, Dongxu and Xiong, Caiming and Hoi, Steven C.H.},
  journal={arXiv preprint arXiv:2201.12086},
  year={2022}
}

@article{wang2017chestx,
  title={ChestX-ray8: Hospital-scale chest X-ray database and benchmarks on weakly-supervised classification and localization of common thorax diseases},
  author={Wang, Xiaosong and Peng, Yifan and Lu, Le and Lu, Zhiyong and Bagheri, Mohammad and Summers, Ronald M},
  journal={Proceedings of the IEEE conference on computer vision and pattern recognition (CVPR)},
  year={2017},
  pages={2097--2106}
}

@article{radford2021learning,
  title={Learning transferable visual models from natural language supervision},
  author={Radford, Alec and Kim, Jong Wook and Hallacy, Chris and others},
  journal={arXiv preprint arXiv:2103.00020},
  year={2021}
}

@misc{maji2013finegrainedvisualclassificationaircraft,
      title={Fine-Grained Visual Classification of Aircraft},
      author={Subhransu Maji and Esa Rahtu and Juho Kannala and Matthew Blaschko and Andrea Vedaldi},
      year={2013},
      eprint={1306.5151},
      archivePrefix={arXiv},
      primaryClass={cs.CV},
      url={https://arxiv.org/abs/1306.5151},
}

@misc{openvino_otx,
  author       = {{Open Edge Platform (Intel)}},
  title        = {OpenVINO™ Training Extensions (OTX)},
  year         = {2025},
  howpublished = {\url{https://github.com/open-edge-platform/training_extensions}},
  note         = {Accessed: 2025-10-13},
}

@inproceedings{liu2022focus,
  title={Where to focus: Investigating hierarchical attention relationship for fine-grained visual classification},
  author={Liu, Yang and Zhou, Lei and Zhang, Pengcheng and Bai, Xiao and Gu, Lin and Yu, Xiaohan and Zhou, Jun and Hancock, Edwin R},
  booktitle={European Conference on Computer Vision},
  pages={57--73},
  year={2022},
  organization={Springer}
}

@inproceedings{wang2023consistency,
  title={Consistency-aware feature learning for hierarchical fine-grained visual classification},
  author={Wang, Rui and Zou, Cong and Zhang, Weizhong and Zhu, Zixuan and Jing, Lihua},
  booktitle={Proceedings of the 31st ACM International Conference on Multimedia},
  pages={2326--2334},
  year={2023}
}

@misc{park2025visuallyconsistenthierarchicalimage,
      title={Visually Consistent Hierarchical Image Classification},
      author={Seulki Park and Youren Zhang and Stella X. Yu and Sara Beery and Jonathan Huang},
      year={2025},
      eprint={2406.11608},
      archivePrefix={arXiv},
      primaryClass={cs.CV},
      url={https://arxiv.org/abs/2406.11608},
}

@misc{tan2021efficientnetv2smallermodelsfaster,
      title={EfficientNetV2: Smaller Models and Faster Training},
      author={Mingxing Tan and Quoc V. Le},
      year={2021},
      eprint={2104.00298},
      archivePrefix={arXiv},
      primaryClass={cs.CV},
      url={https://arxiv.org/abs/2104.00298},
}

@InProceedings{pmlr-v139-touvron21a,
  title =     {Training data-efficient image transformers \& distillation through attention},
  author =    {Touvron, Hugo and Cord, Matthieu and Douze, Matthijs and Massa, Francisco and Sablayrolles, Alexandre and Jegou, Herve},
  booktitle = {International Conference on Machine Learning},
  pages =     {10347--10357},
  year =      {2021},
  volume =    {139},
  month =     {July}
}

@misc{adamw,
      title={Decoupled Weight Decay Regularization},
      author={Ilya Loshchilov and Frank Hutter},
      year={2019},
      eprint={1711.05101},
      archivePrefix={arXiv},
      primaryClass={cs.LG},
      url={https://arxiv.org/abs/1711.05101},
}

@inproceedings{optuna,
author = {Akiba, Takuya and Sano, Shohei and Yanase, Takuma and Ohta, Takeru and Koyama, Masanori},
title = {Optuna: A Next-generation Hyperparameter Optimization Framework},
booktitle = {Proceedings of the 25th ACM SIGKDD International Conference on Knowledge Discovery \& Data Mining},
pages = {2623--2631},
year = {2019},
publisher = {ACM}
}

@article{JMLR:v9:vandermaaten08a,
  author  = {Laurens van der Maaten and Geoffrey Hinton},
  title   = {Visualizing Data using t-SNE},
  journal = {Journal of Machine Learning Research},
  year    = {2008},
  volume  = {9},
  number  = {86},
  pages   = {2579--2605},
  url     = {http://jmlr.org/papers/v9/vandermaaten08a.html}
}

@misc{hinton2015,
      title={Distilling the Knowledge in a Neural Network}, 
      author={Geoffrey Hinton and Oriol Vinyals and Jeff Dean},
      year={2015},
      eprint={1503.02531},
      archivePrefix={arXiv},
      primaryClass={stat.ML},
      url={https://arxiv.org/abs/1503.02531}, 
}

@article{bjerge2023hierarchical,
  title={Hierarchical classification of insects with multitask learning and anomaly detection},
  author={Bjerge, Kim and Geissmann, Quentin and Alison, Jamie and Mann, Hjalte MR and H{\o}ye, Toke T and Dyrmann, Mads and Karstoft, Henrik},
  journal={Ecological Informatics},
  volume={77},
  pages={102278},
  year={2023},
  publisher={Elsevier}
}

@article{elhamod2022hierarchy,
  title={Hierarchy-guided neural network for species classification},
  author={Elhamod, Mohannad and Diamond, Kelly M and Maga, A Murat and Bakis, Yasin and Bart Jr, Henry L and Mabee, Paula and Dahdul, Wasila and Leipzig, Jeremy and Greenberg, Jane and Avants, Brian and others},
  journal={Methods in Ecology and Evolution},
  volume={13},
  number={3},
  pages={642--652},
  year={2022},
  publisher={Wiley Online Library}
}

@article{yu2025hierarchical,
  title={Hierarchical skin lesion image classification with prototypical decision tree},
  author={Yu, Zhen and Nguyen, Toan D and Ju, Lie and Gal, Yaniv and Sashindranath, Maithili and Bonnington, Paul and Zhang, Lei and Mar, Victoria and Ge, Zongyuan},
  journal={npj Digital Medicine},
  volume={8},
  number={1},
  pages={26},
  year={2025},
  publisher={Nature Publishing Group UK London}
}

@inproceedings{sali2020hierarchical,
  title={Hierarchical deep convolutional neural networks for multi-category diagnosis of gastrointestinal disorders on histopathological images},
  author={Sali, Rasoul and Adewole, Sodiq and Ehsan, Lubaina and Denson, Lee A and Kelly, Paul and Amadi, Beatrice C and Holtz, Lori and Ali, Syed Asad and Moore, Sean R and Syed, Sana and others},
  booktitle={2020 IEEE International Conference on Healthcare Informatics (ICHI)},
  pages={1--6},
  year={2020},
  organization={IEEE}
}

@article{zhu2017b,
  title={B-CNN: branch convolutional neural network for hierarchical classification},
  author={Zhu, Xinqi and Bain, Michael},
  journal={arXiv preprint arXiv:1709.09890},
  year={2017}
}

@inproceedings{li2021mmf,
  title={MMF: multi-task multi-structure fusion for hierarchical image classification},
  author={Li, Xiaoni and Zhou, Yucan and Zhou, Yu and Wang, Weiping},
  booktitle={International Conference on Artificial Neural Networks},
  pages={61--73},
  year={2021},
  organization={Springer}
}

@article{valmadre2022hierarchical,
  title={Hierarchical classification at multiple operating points},
  author={Valmadre, Jack},
  journal={Advances in Neural Information Processing Systems},
  volume={35},
  pages={18034--18045},
  year={2022}
}

@inproceedings{chen2018fine,
  title={Fine-grained representation learning and recognition by exploiting hierarchical semantic embedding},
  author={Chen, Tianshui and Wu, Wenxi and Gao, Yuefang and Dong, Le and Luo, Xiaonan and Lin, Liang},
  booktitle={Proceedings of the 26th ACM international conference on Multimedia},
  pages={2023--2031},
  year={2018}
}

@article{yu2022coca,
  title={Coca: Contrastive captioners are image-text foundation models},
  author={Yu, Jiahui and Wang, Zirui and Vasudevan, Vijay and Yeung, Legg and Seyedhosseini, Mojtaba and Wu, Yonghui},
  journal={arXiv preprint arXiv:2205.01917},
  year={2022}
}

@article{li2021align,
  title={Align before fuse: Vision and language representation learning with momentum distillation},
  author={Li, Junnan and Selvaraju, Ramprasaath and Gotmare, Akhilesh and Joty, Shafiq and Xiong, Caiming and Hoi, Steven Chu Hong},
  journal={Advances in neural information processing systems},
  volume={34},
  pages={9694--9705},
  year={2021}
}

@inproceedings{yang2022vision,
  title={Vision-language pre-training with triple contrastive learning},
  author={Yang, Jinyu and Duan, Jiali and Tran, Son and Xu, Yi and Chanda, Sampath and Chen, Liqun and Zeng, Belinda and Chilimbi, Trishul and Huang, Junzhou},
  booktitle={Proceedings of the IEEE/CVF conference on computer vision and pattern recognition},
  pages={15671--15680},
  year={2022}
}

@article{lester2021power,
  title={The power of scale for parameter-efficient prompt tuning},
  author={Lester, Brian and Al-Rfou, Rami and Constant, Noah},
  journal={arXiv preprint arXiv:2104.08691},
  year={2021}
}

@article{hu2025multi,
  title={Multi-Scale Attention-Driven Hierarchical Learning for Fine-Grained Visual Categorization},
  author={Hu, Zhihuai and Kojima, Rihito and Han, Xian-Hua},
  journal={Electronics},
  volume={14},
  number={14},
  pages={2869},
  year={2025},
  publisher={MDPI}
}

@article{mumtaz2022hierarchy,
  title={Hierarchy-based semantic embeddings for single-valued \& multi-valued categorical variables},
  author={Mumtaz, Summaya and Giese, Martin},
  journal={Journal of Intelligent Information Systems},
  volume={58},
  number={3},
  pages={613--640},
  year={2022},
  publisher={Springer}
}

@inproceedings{chang2021your,
  title={Your" flamingo" is my" bird": Fine-grained, or not},
  author={Chang, Dongliang and Pang, Kaiyue and Zheng, Yixiao and Ma, Zhanyu and Song, Yi-Zhe and Guo, Jun},
  booktitle={Proceedings of the IEEE/CVF conference on computer vision and pattern recognition},
  pages={11476--11485},
  year={2021}
}

@inproceedings{hu2023vl,
  title={Vl-pet: Vision-and-language parameter-efficient tuning via granularity control},
  author={Hu, Zi-Yuan and Li, Yanyang and Lyu, Michael R and Wang, Liwei},
  booktitle={Proceedings of the IEEE/CVF International Conference on Computer Vision},
  pages={3010--3020},
  year={2023}
}

@article{liu2023parameter,
  title={Parameter-efficient transfer learning for medical visual question answering},
  author={Liu, Jiaxiang and Hu, Tianxiang and Zhang, Yan and Feng, Yang and Hao, Jin and Lv, Junhui and Liu, Zuozhu},
  journal={IEEE Transactions on Emerging Topics in Computational Intelligence},
  volume={8},
  number={4},
  pages={2816--2826},
  year={2023},
  publisher={IEEE}
}

@inproceedings{van2023open,
  title={Open-ended medical visual question answering through prefix tuning of language models},
  author={Van Sonsbeek, Tom and Derakhshani, Mohammad Mahdi and Najdenkoska, Ivona and Snoek, Cees GM and Worring, Marcel},
  booktitle={International Conference on Medical Image Computing and Computer-Assisted Intervention},
  pages={726--736},
  year={2023},
  organization={Springer}
}

@inproceedings{zhang2024cream,
  title={CREAM: coarse-to-fine retrieval and multi-modal efficient tuning for document VQA},
  author={Zhang, Jinxu and Yu, Yongqi and Zhang, Yu},
  booktitle={Proceedings of the 32nd ACM International Conference on Multimedia},
  pages={925--934},
  year={2024}
}

@inproceedings{duan2024cityllava,
  title={Cityllava: Efficient fine-tuning for vlms in city scenario},
  author={Duan, Zhizhao and Cheng, Hao and Xu, Duo and Wu, Xi and Zhang, Xiangxie and Ye, Xi and Xie, Zhen},
  booktitle={Proceedings of the IEEE/CVF Conference on Computer Vision and Pattern Recognition},
  pages={7180--7189},
  year={2024}
}

@inproceedings{wang2023efficientvlm,
  title={Efficientvlm: Fast and accurate vision-language models via knowledge distillation and modal-adaptive pruning},
  author={Wang, Tiannan and Zhou, Wangchunshu and Zeng, Yan and Zhang, Xinsong},
  booktitle={Findings of the association for computational linguistics: ACL 2023},
  pages={13899--13913},
  year={2023}
}

@article{lu2023uniadapter,
  title={Uniadapter: Unified parameter-efficient transfer learning for cross-modal modeling},
  author={Lu, Haoyu and Huo, Yuqi and Yang, Guoxing and Lu, Zhiwu and Zhan, Wei and Tomizuka, Masayoshi and Ding, Mingyu},
  journal={arXiv preprint arXiv:2302.06605},
  year={2023}
}

@inproceedings{yang2024cross,
  title={Cross-modal adapter: Parameter-efficient transfer learning approach for vision-language models},
  author={Yang, Juncheng and Li, Zuchao and Xie, Shuai and Zhu, Weiping and Yu, Wei and Li, Shijun},
  booktitle={2024 IEEE International Conference on Multimedia and Expo (ICME)},
  pages={1--6},
  year={2024},
  organization={IEEE}
}

@article{madan2024revisiting,
  title={Revisiting few-shot object detection with vision-language models},
  author={Madan, Anish and Peri, Neehar and Kong, Shu and Ramanan, Deva},
  journal={Advances in Neural Information Processing Systems},
  volume={37},
  pages={19547--19560},
  year={2024}
}

@article{sharshar2025vision,
  title={Vision-language models for edge networks: A comprehensive survey},
  author={Sharshar, Ahmed and Khan, Latif U and Ullah, Waseem and Guizani, Mohsen},
  journal={IEEE Internet of Things Journal},
  year={2025},
  publisher={IEEE}
}
